\documentclass[10pt,twocolumn,letterpaper]{article}

\usepackage{cvpr}
\usepackage{times}
\usepackage{epsfig}
\usepackage{graphicx}
\usepackage{amsmath}
\usepackage{amssymb}
\usepackage{subcaption}
\usepackage{fixltx2e}
\usepackage{multirow}

\newcommand\blfootnote[1]{%
  \begingroup
  \renewcommand\thefootnote{}\footnote{#1}%
  \addtocounter{footnote}{-1}%
  \endgroup
}

\usepackage[pagebackref=true,breaklinks=true,letterpaper=true,colorlinks,bookmarks=false]{hyperref}

\cvprfinalcopy 

\def\cvprPaperID{156} 

\ifcvprfinal\fi
\begin{document}

\title{Eye Tracking for Everyone}

\author{Kyle Krafka$^{*\dagger}$\hspace*{5mm} Aditya Khosla$^{*\ddagger}$\hspace*{5mm} Petr Kellnhofer$^{\ddagger\star}$\hspace*{5mm} Harini Kannan$^{\ddagger}$\\Suchendra Bhandarkar$^{\dagger}$\hspace*{6mm} Wojciech Matusik$^{\ddagger}$\hspace*{6mm} Antonio Torralba$^{\ddagger}$\\
$^{\dagger}$University of Georgia\hspace*{5mm}$^{\ddagger}$Massachusetts Institute of Technology\hspace*{5mm}$^{\star}$MPI Informatik\\
{\tt\small \{krafka, suchi\}@cs.uga.edu}\hspace*{4mm}{\tt\small \{khosla, pkellnho, hkannan, wojciech, torralba\}@csail.mit.edu}
}

\maketitle

\begin{abstract}
From scientific research to commercial applications, eye tracking is an important tool across many domains. Despite its range of applications, eye tracking has yet to become a pervasive technology. We believe that we can put the power of eye tracking in everyone's palm by building eye tracking software that works on commodity hardware such as mobile phones and tablets, without the need for additional sensors or devices.
We tackle this problem by introducing \emph{GazeCapture}, the first large-scale dataset for eye tracking, containing data from over 1450 people consisting of almost $2.5M$ frames. Using GazeCapture, we train \emph{iTracker}, a convolutional neural network for eye tracking, which achieves a significant reduction in error over previous approaches while running in real time (10--15fps) on a modern mobile device. Our model achieves a prediction error of 1.71cm and 2.53cm without calibration on mobile phones and tablets respectively. With calibration, this is reduced to 1.34cm and 2.12cm. Further, we demonstrate that the features learned by iTracker generalize well to other datasets, achieving state-of-the-art results. The code, data, and models are available at \href{http://gazecapture.csail.mit.edu}{http://gazecapture.csail.mit.edu}.

\blfootnote{$^*$ indicates equal contribution\\Corresponding author: Aditya Khosla (\href{mailto:khosla@csail.mit.edu}{khosla@csail.mit.edu})}
\end{abstract}

\section{Introduction}

\begin{figure}[ht]
	\centering
	\includegraphics*[clip, width = \linewidth]{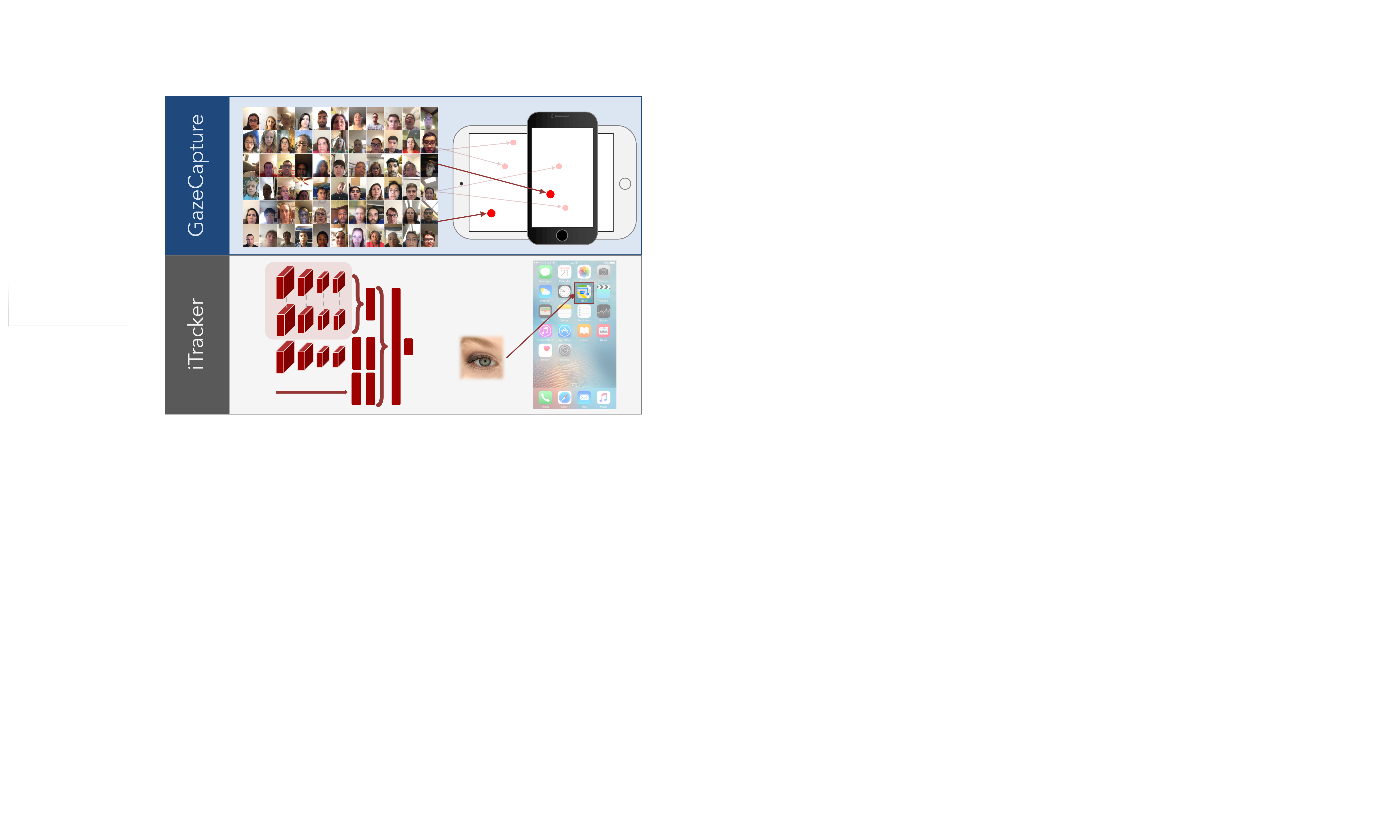}
	\caption{In this work, we develop GazeCapture, the first large-scale eye tracking dataset captured via crowdsourcing. Using GazeCapture, we train iTracker, a convolutional neural network for robust gaze prediction.}
	\label{fig:Teaser}
\end{figure}

From human--computer interaction techniques~\cite{jacob2003eye,majaranta2014eye,morimoto2005eye} to
medical diagnoses~\cite{holzman1974eye} to psychological studies~\cite{rayner1998eye} to computer vision~\cite{borji2013state,karthikeyan2013and}, eye tracking
has applications in many areas~\cite{duchowski2007eye}. Gaze is the externally-observable indicator of
human visual attention, and many have attempted to record it, dating back to
the late eighteenth century~\cite{huey1908psychology}. Today, a variety of solutions exist (many of them
commercial) but all suffer from one or more of the following: high cost (\eg, Tobii X2-60), custom
or invasive hardware (\eg, Eye Tribe, Tobii EyeX) or inaccuracy under real-world conditions (\eg,~\cite{mora2014eyediap,sugano2014learning,zhang15_cvpr}).
These factors prevent eye tracking from becoming a pervasive technology that should be available to anyone with a reasonable camera (\eg, a smartphone or a webcam).
In this work, our goal is to overcome these challenges to bring eye tracking to everyone.

We believe that this goal can be achieved by developing systems that work reliably on mobile devices such as smartphones and tablets, without the need for any external attachments (Fig.~\ref{fig:Teaser}). Mobile devices offer several benefits over other platforms: (1) widespread use---more than a third of the world's population is estimated to have smartphones by 2019~\cite{statista}, far exceeding the number of desktop/laptop users; (2) high adoption rate of technology upgrades---a large proportion of people have the latest hardware allowing for the use of computationally expensive methods, such as convolutional neural networks (CNNs), in real-time; (3) the heavy usage of cameras on mobile devices has lead to rapid development and deployment of camera technology, and (4) the fixed position of the camera relative to the screen reduces the number of unknown parameters, potentially allowing for the development of high-accuracy calibration-free tracking.

The recent success of deep learning has been apparent in a variety of domains in computer
vision~\cite{krizhevsky2012imagenet,girshick2014rich,taigman2014deepface,nips15_recasens,iccv15_khosla}, but
its impact on improving the performance of eye tracking has been rather limited~\cite{zhang15_cvpr}.
We believe that this is due to the lack of availability of large-scale data, with the largest datasets having $\sim$50 subjects~\cite{huang2015tabletgaze,sugano2014learning}.
In this work, using crowdsourcing, we build \emph{GazeCapture}, a mobile-based eye tracking dataset containing almost 1500 subjects
from a wide variety of backgrounds, recorded under variable lighting conditions
and unconstrained head motion.

Using GazeCapture, we train \emph{iTracker}, a convolutional neural network (CNN) learned end-to-end for gaze prediction. iTracker does not rely on any preexisting
systems for head pose estimation or other manually-engineered features for prediction. Training the network with just crops of both eyes and the face, we outperform
existing eye tracking approaches in this domain by a significant margin. While our network achieves state-of-the-art performance in terms of accuracy, the size of the
inputs and number of parameters make it difficult to use in real-time on a mobile device. To address this we apply ideas from the work on dark knowledge by Hinton \etal~\cite{hinton2015distilling}
to train a smaller and faster network that achieves real-time performance on mobile devices with a minimal loss in accuracy.

Overall, we take a significant step towards putting the power of eye tracking in everyone's palm.

\section{Related Work}
There has been a plethora of work on predicting gaze. Here, we give a brief overview of some of the existing gaze estimation methods and urge the reader to look at this excellent survey paper~\cite{hansen2010eye} for a more complete picture. We also discuss the differences between GazeCapture and other popular gaze estimation datasets.

\textbf{Gaze estimation:} Gaze estimation methods can be divided into model-based or appearance-based~\cite{hansen2010eye}. Model-based approaches use a geometric model of an eye and can be subdivided into corneal-reflection-based and shape-based methods. Corneal-reflection-based methods~\cite{yoo2005novel,zhu2005eye,zhu2006nonlinear,hennessey2006single} rely on external light sources to detect eye features. On the other hand, shape-based methods~\cite{ishikawa2004passive,chen20083d,valenti2012combining,hansen2005eye} infer gaze direction from observed eye shapes, such as pupil centers and iris edges. These approaches tend to suffer with low image quality and variable lighting conditions, as in our scenario. Appearance-based methods~\cite{tan2002appearance,sewell2010real,lu2014adaptive,lu2014learning,torricelli2008neural,baluja1994non} directly use eyes as input and can potentially work on low-resolution images. Appearance-based methods are believed~\cite{zhang15_cvpr} to require larger amounts of user-specific training data as compared to model-based methods. However, we show that our model is able to generalize well to novel faces without needing user-specific data. While calibration is helpful, its impact is not as significant as in other approaches given our model's inherent generalization ability achieved through the use of deep learning and large-scale data. Thus, our model does not have to rely on visual saliency maps~\cite{chen2011probabilistic,sugano2013appearance} or key presses~\cite{sugano2008incremental} to achieve accurate calibration-free gaze estimation. Overall, iTracker is a data-driven appearance-based model learned end-to-end without using any hand-engineered features such as head pose or eye center location. We also demonstrate that our trained networks can produce excellent features for gaze prediction (that outperform hand-engineered features) on other datasets despite not having been trained on them.

\textbf{Gaze datasets:} There are a number of publicly available gaze datasets in the community~\cite{mcmurrough2012eye,weidenbacher2007comprehensive,smith2013gaze,mora2014eyediap,sugano2014learning,zhang15_cvpr,huang2015tabletgaze}. We summarize the distinctions from these datasets in Tbl.~\ref{tbl:datasets}. Many of the earlier datasets~\cite{mcmurrough2012eye,weidenbacher2007comprehensive,smith2013gaze} do not contain significant variation in head pose or have a coarse gaze point sampling density. We overcome this by encouraging participants to move their head while recording and generating a random distribution of gaze points for each participant. While some of the modern datasets follow a similar approach~\cite{sugano2014learning,mora2014eyediap,zhang15_cvpr,huang2015tabletgaze}, their scale---especially in the number of participants---is rather limited. We overcome this through the use of crowdsourcing, allowing us to build a dataset with $\sim$30 times as many participants as the current largest dataset. Further, unlike~\cite{zhang15_cvpr}, given our recording permissions, we can release the complete images without post-processing. We believe that GazeCapture will serve as an invaluable resource for future work in this domain.

\begin{table}
 \centering
  \small
 \begin{tabular}{cccccc}
 \hline
  & \# People & Poses & Targets & Illum. & Images\\\hline\hline
  ~\cite{mcmurrough2012eye} & 20 & 1 & 16 & 1 & videos\\\hline
  ~\cite{weidenbacher2007comprehensive} & 20 & 19 & 2--9 & 1 & 1,236\\\hline
  ~\cite{smith2013gaze} & 56 & 5 & 21 & 1 & 5,880\\\hline
  ~\cite{mora2014eyediap} & 16 & cont. & cont. & 2 & videos\\\hline
  ~\cite{sugano2014learning} & 50 & 8+synth. & 160 & 1 & 64,000\\\hline
  ~\cite{zhang15_cvpr} & 15 & cont. & cont. & cont. & 213,659\\\hline
  ~\cite{huang2015tabletgaze} & 51 & cont. & 35 & cont. & videos\\\hline\hline
  \textbf{Ours} & \textbf{1474} & \textbf{cont.} & \textbf{13+cont.} & \textbf{cont.} & \textbf{2,445,504} \\\hline
 \end{tabular}
 \caption{Comparison of our GazeCapture dataset with popular publicly available datasets. GazeCapture has approximately $30$ times as many participants and $10$ times as many frames as the largest datasets and contains a significant amount of variation in pose and illumination, as it was recorded using crowdsourcing. We use the following abbreviations: \textit{cont.} for continuous, \textit{illum.} for illumination, and \textit{synth.} for synthesized.}
 \label{tbl:datasets}
\end{table}

\section{GazeCapture: A Large-Scale Dataset}

In this section, we describe how we achieve our goal of scaling up the collection of eye tracking data. We find that most existing eye tracking datasets have been collected by researchers inviting participants to the lab, a process that leads to a lack of variation in the data and is costly and inefficient to scale up. We overcome these limitations through the use of crowdsourcing, a popular approach for collecting large-scale datasets~\cite{russakovsky2014imagenet,iccv15_khosla,places2,nips15_recasens}. In Sec.~\ref{sec:crowdsource}, we describe the process of obtaining reliable data via crowdsourcing and in Sec.~\ref{sec:characteristics}, we compare the characteristics of GazeCapture with existing datasets.

\subsection{Collecting Eye Tracking Data}
\label{sec:crowdsource}
Our goal here is to develop an approach for collecting eye tracking data on mobile devices that is (1) scalable, (2) reliable, and (3) produces large variability. Below, we describe, in detail, how we achieve each of these three goals.

\textbf{Scalability:} In order for our approach to be scalable, we must design an automated mechanism for gathering data and reaching participants. Crowdsourcing is a popular technique researchers use to achieve scalability. The primary difficulty with this approach is that most crowdsourcing platforms are designed to be used on laptops/desktops and provide limited flexibility required to design the desired user experience. Thus, we decided to use a hybrid approach, combining the scalable workforce of crowdsourcing platforms together with the design freedom provided by building custom mobile applications. Specifically, we built an iOS application, also named GazeCapture\footnote{\href{http://apple.co/1q1Ozsg}{http://apple.co/1q1Ozsg}}, capable of recording and uploading gaze tracking data, and used Amazon Mechanical Turk (AMT) as a platform for recruiting people to use our application. On AMT, the workers were provided detailed instructions on how to download the application from Apple's App Store and complete the task.

We chose to build the GazeCapture application for Apple's iOS because of the large-scale adoption of latest Apple devices, and the ease of deployment across multiple device types such as iPhones and iPads using a common code base. Further, the lack of fragmentation in the versions of the operating system (as compared to other platforms) significantly simplified the development process. Additionally, we released the application publicly to the App Store (as opposed to a beta release with limited reach) simplifying installation of our application, thereby further aiding the scalability of our approach.

\textbf{Reliability:} The simplest rendition of our GazeCapture application could involve showing workers dots on a screen at random locations and recording their gaze using the front-facing camera. While this approach may work well when calling individual participants to the lab, it is not likely to produce reliable results without human supervision. Thus, we must design an automatic mechanism that ensures workers are paying attention and fixating directly on the dots shown on the screen.

\begin{figure}[t]
\begin{center}
   \includegraphics[width=\linewidth]{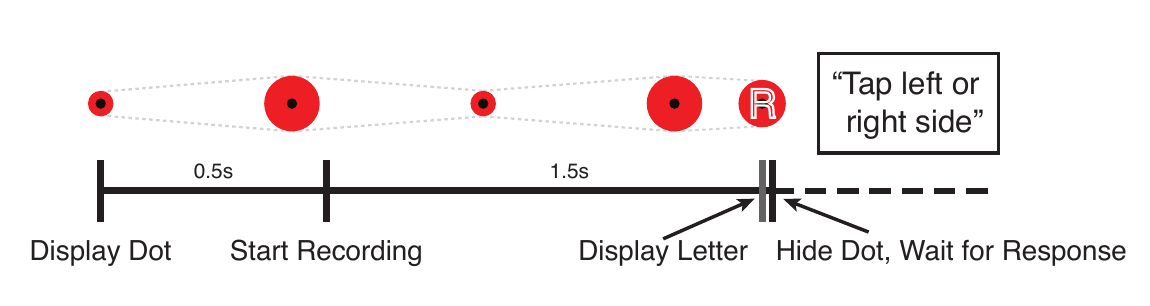}
\end{center}
\vspace*{-6mm}
\caption{The timeline of the display of an individual dot. Dotted gray lines
indicate how the dot changes size over time to keep attention.}
\label{fig:timeline}
\end{figure}

First, to avoid distraction from notifications, we ensure that the worker uses \textit{Airplane Mode} with no network connection throughout the task, until the task is complete and ready to be uploaded. Second, instead of showing a plain dot, we show a pulsating red circle around the dot, as shown in Fig.~\ref{fig:timeline}, that directs the fixation of the eye to lie in the middle of that circle. This pulsating dot is shown for approximately $2$s and we start the recording $0.5$sec. after the dot moves to a new location to allow enough time for the worker to fixate at the dot location. Third, towards the end of the $2$sec. window, a small letter, \textit{L} or \textit{R} is displayed for $0.05$sec.---based on this letter, the worker is required to tap either the left (\textit{L}) or right (\textit{R}) side of the screen. This serves as a means to monitor the worker's attention and provide engagement with the application. If the worker taps the wrong side, they are warned and must repeat the dot again. Last, we use the real-time face detector built into iOS to ensure that the worker's face is visible in a large proportion of the recorded frames. This is critical as we cannot hope to track where someone is looking without a picture of their eyes.

\textbf{Variability:} In order to learn a robust eye tracking model, significant variability in the data is important. We believe that this variability is critical to achieving high-accuracy calibration-free eye tracking. Thus, we designed our setup to explicitly encourage high variability.

First, given our use of crowdsourcing, we expect to have a large variability in pose, appearance, and illumination. Second, to encourage further variability in pose, we tell the workers to continuously move their head and the distance of the phone relative to them by showing them an instructional video with a person doing the same. Last, we force workers to change the orientation of their mobile device after every 60 dots. This change can be detected using the built-in sensors on the device. This changes the relative position of the camera and the screen providing further variability.

\begin{figure*}[t]
\begin{center}
   \includegraphics[width=1.0\linewidth]{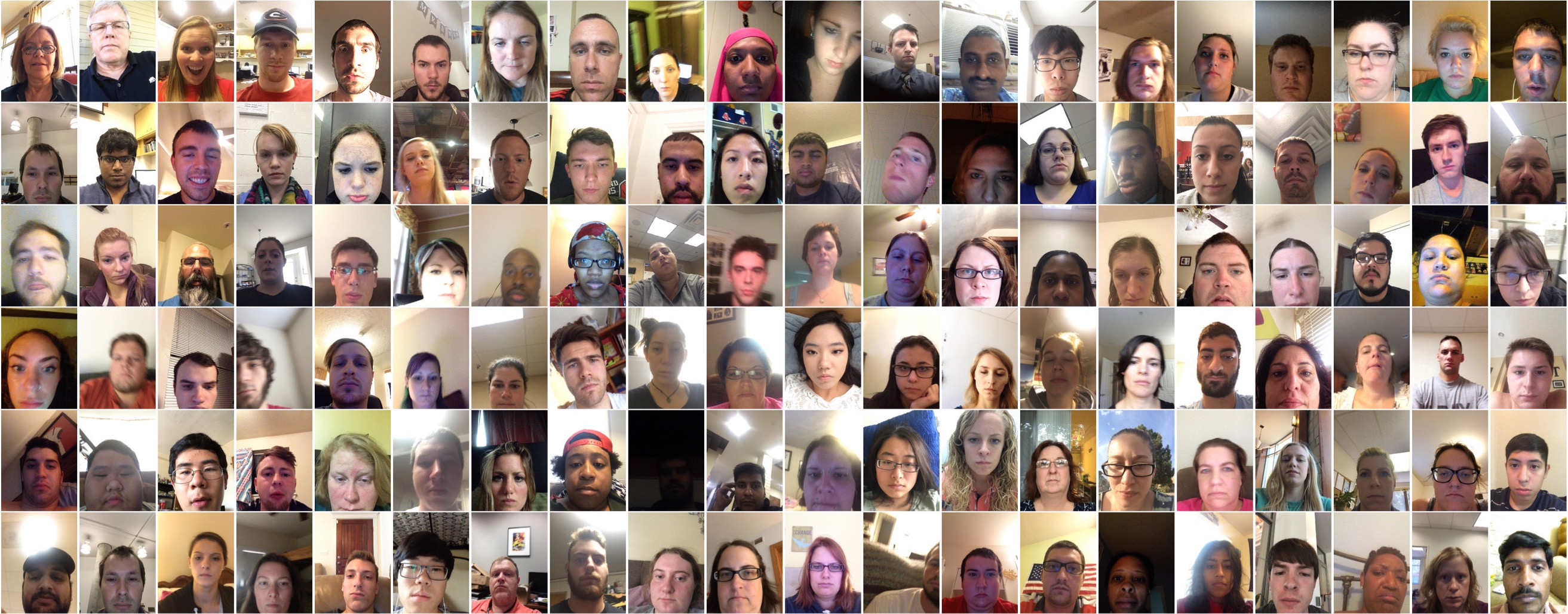}
   \caption{Sample frames from our GazeCapture dataset. Note the significant variation in illumination, head pose, appearance, and background. This variation allows us to learn robust models that generalize well to novel faces.}
  \label{fig:faces}
\end{center}
\end{figure*}

\textbf{Implementation details:} Here, we provide some implementation details that may be helpful for other researchers conducting similar studies. In order to associate each mobile device with an AMT task, we provided each worker with a unique code in AMT that they subsequently typed into their mobile application. The dot locations were both random and from 13 fixed locations (same locations as Fig. 3 of~\cite{xu2015turkergaze})---we use the fixed locations to study the effect of calibration (Sec.~\ref{sec:calibration}). We displayed a total of 60 dots\footnote{This was the number of dots displayed when the user entered a code provided via AMT. When the user did not enter a code (typical case when the application is downloaded directly from the App Store), they were shown 8 dots per orientation to keep them engaged.} for each orientation of the device\footnote{Three orientations for iPhones and four orientations for iPads following their natural use cases.} leading to a task duration of $\sim$10min. Each worker was only allowed to complete the task once and we paid them \$1--\$1.50. We uploaded the data as individual frames rather than a video to avoid compression artifacts. Further, while we did not use it in this work, we also recorded device motion sensor data. We believe that this could be a useful resource for other researchers in the future.

\subsection{Dataset Characteristics}
\label{sec:characteristics}
We collected data from a total of 1474 subjects: 1103 subjects through AMT, 230 subjects through in-class recruitment at UGA, and 141 subjects through other various App Store downloads. This resulted in a total of $2,445,504$ frames with corresponding fixation locations. Sample frames are shown in Fig.~\ref{fig:faces}. 1249 subjects used iPhones while 225 used iPads, resulting in a total of $\sim2.1$M and $\sim360$k frames from each of the devices respectively.

\begin{figure}[ht]
	\centering
	\begin{subfigure}[t]{0.167\textwidth}
		\includegraphics[width=0.985\textwidth]{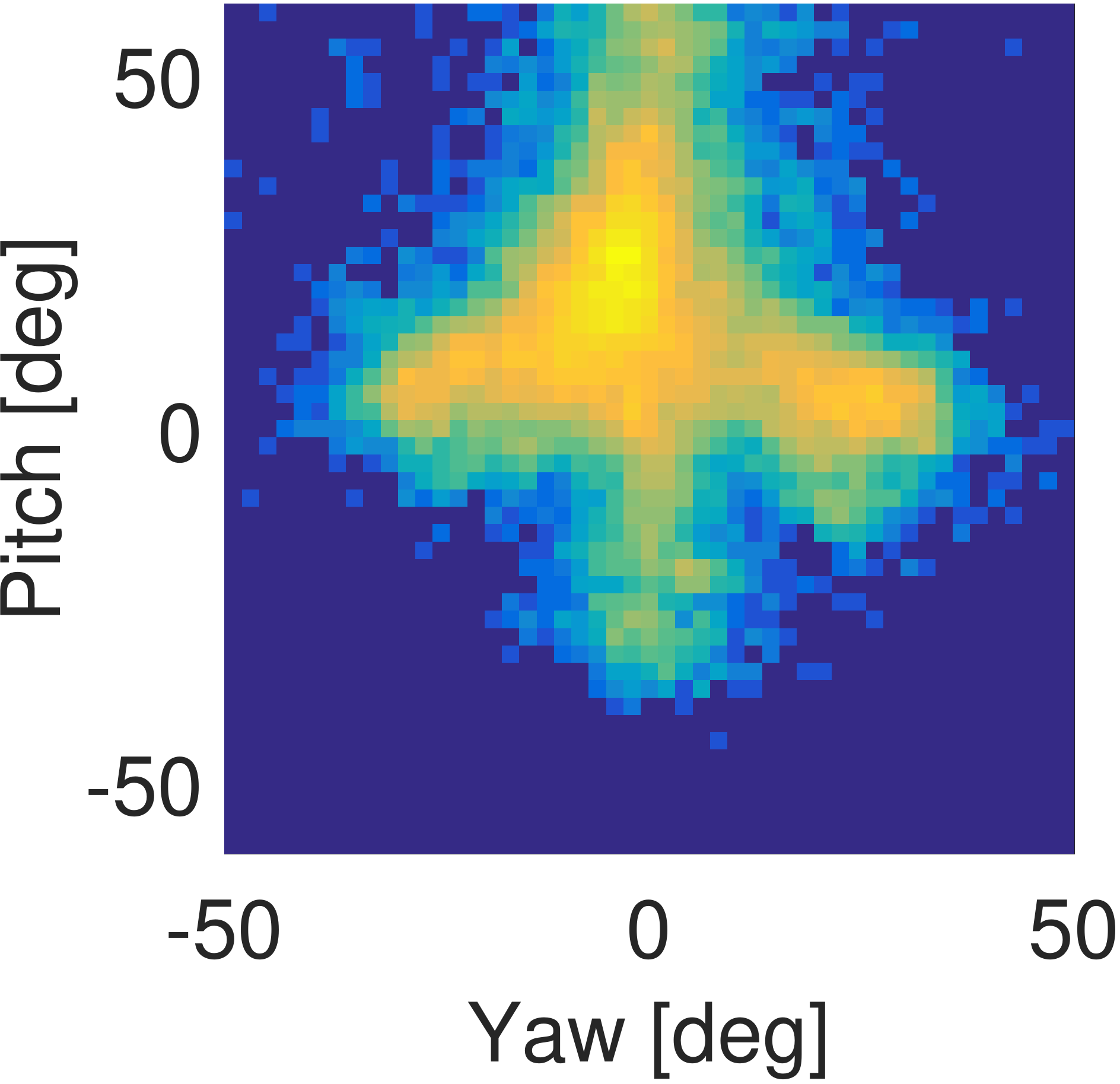}
        \caption{$\mathbf{h}$(TabletGaze)}
    \end{subfigure}%
    \begin{subfigure}[t]{0.167\textwidth}
		\includegraphics[width=0.985\textwidth]{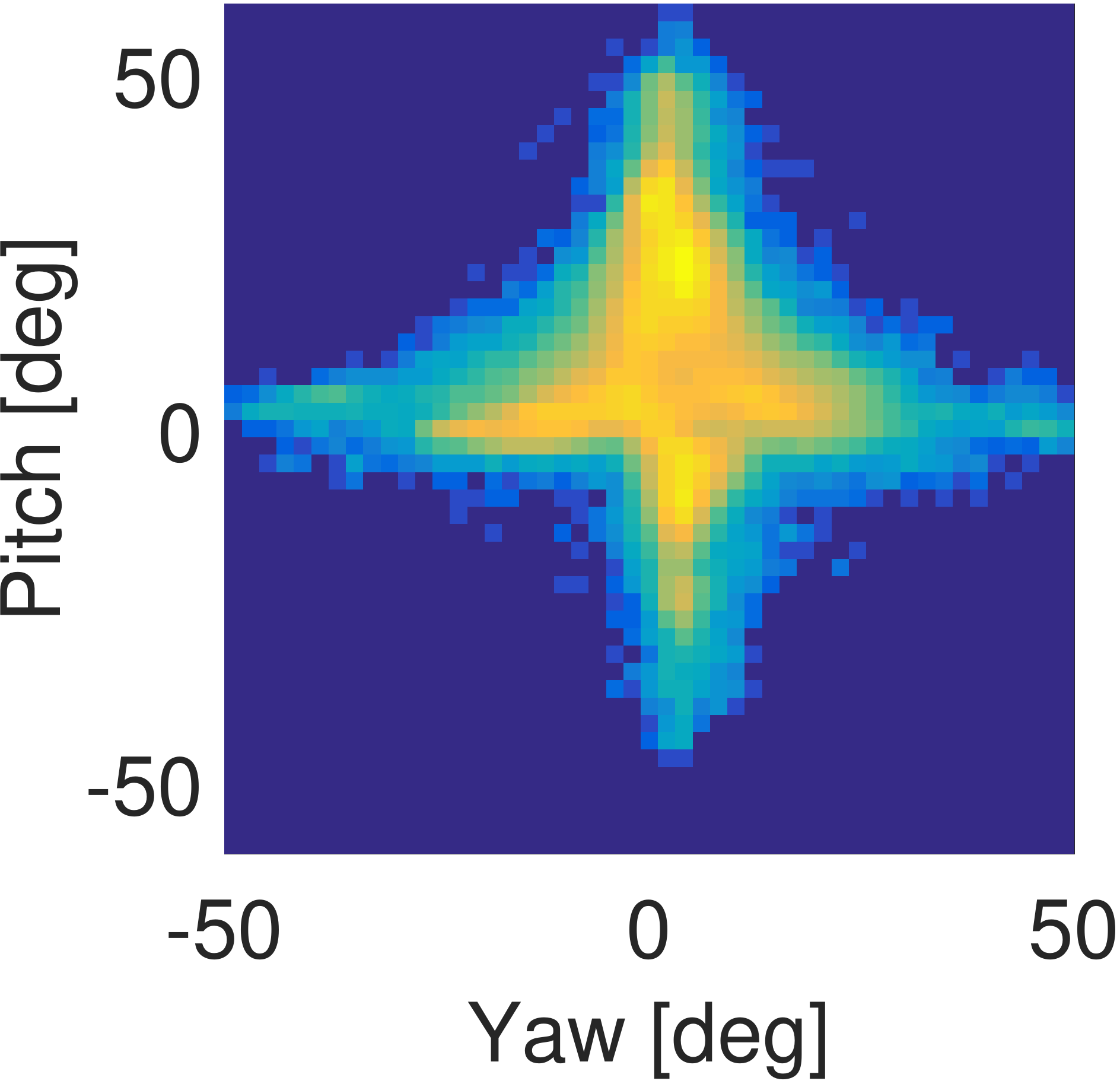}
        \caption{$\mathbf{h}$(MPIIGaze)}
    \end{subfigure}%
    \begin{subfigure}[t]{0.167\textwidth}
		\includegraphics[width=0.985\textwidth]{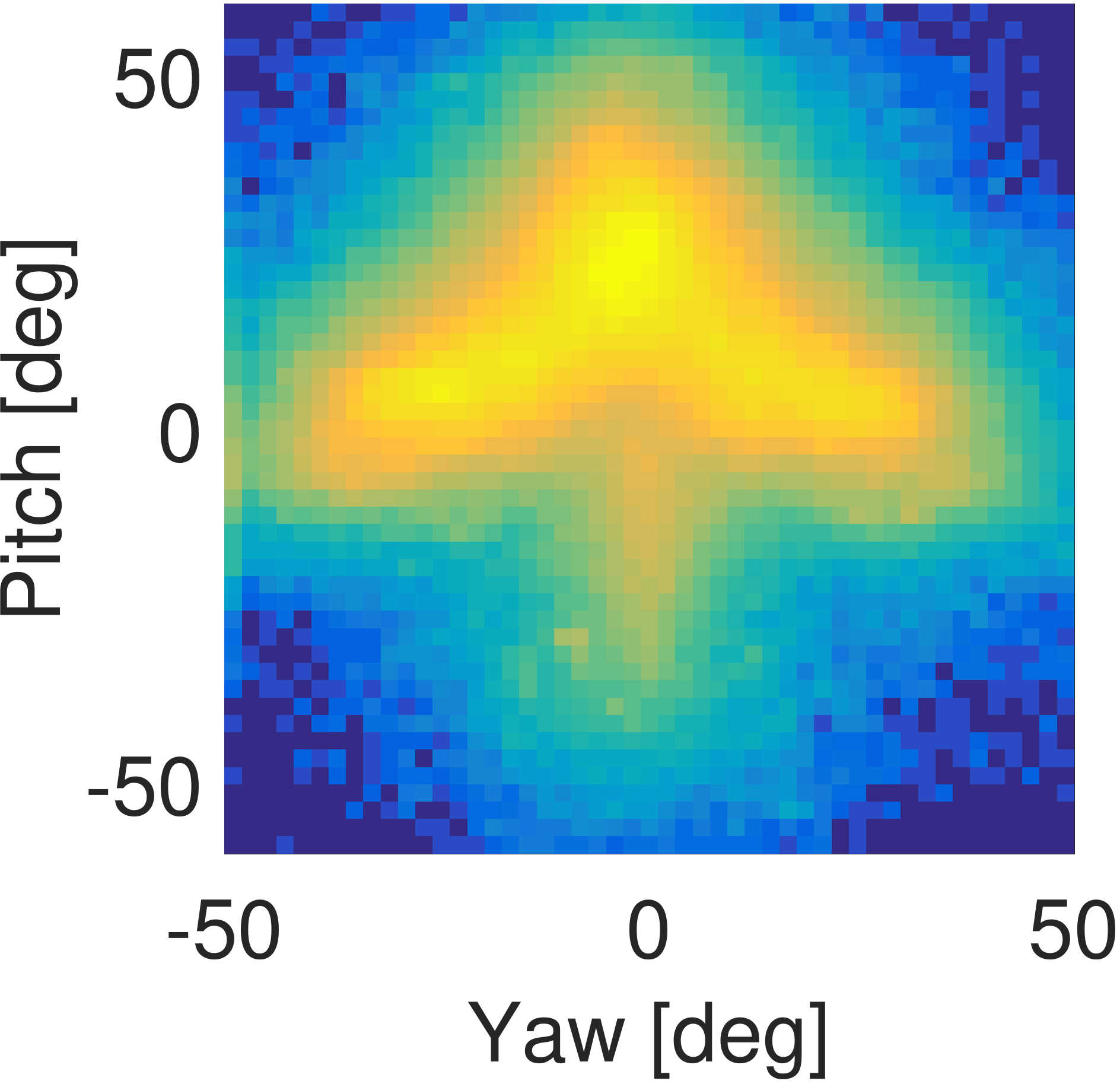}
        \caption{$\mathbf{h}$(GazeCapture)}
    \end{subfigure}%

    \vspace{0.2cm}

	\begin{subfigure}[t]{0.167\textwidth}
		\includegraphics[width=0.985\textwidth]{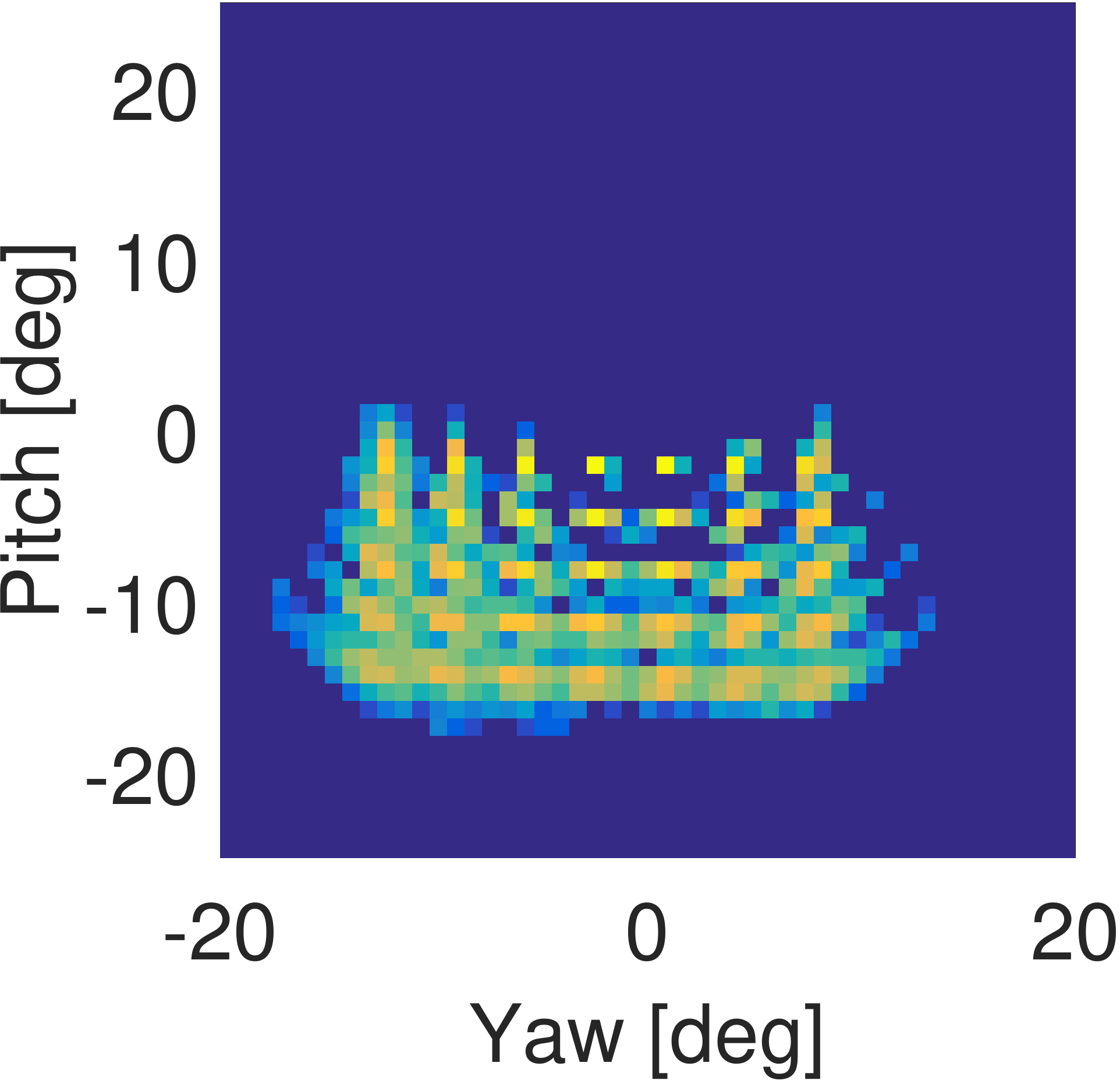}
        \caption{$\mathbf{g}$(TabletGaze)}
    \end{subfigure}%
    \begin{subfigure}[t]{0.167\textwidth}
		\includegraphics[width=0.985\textwidth]{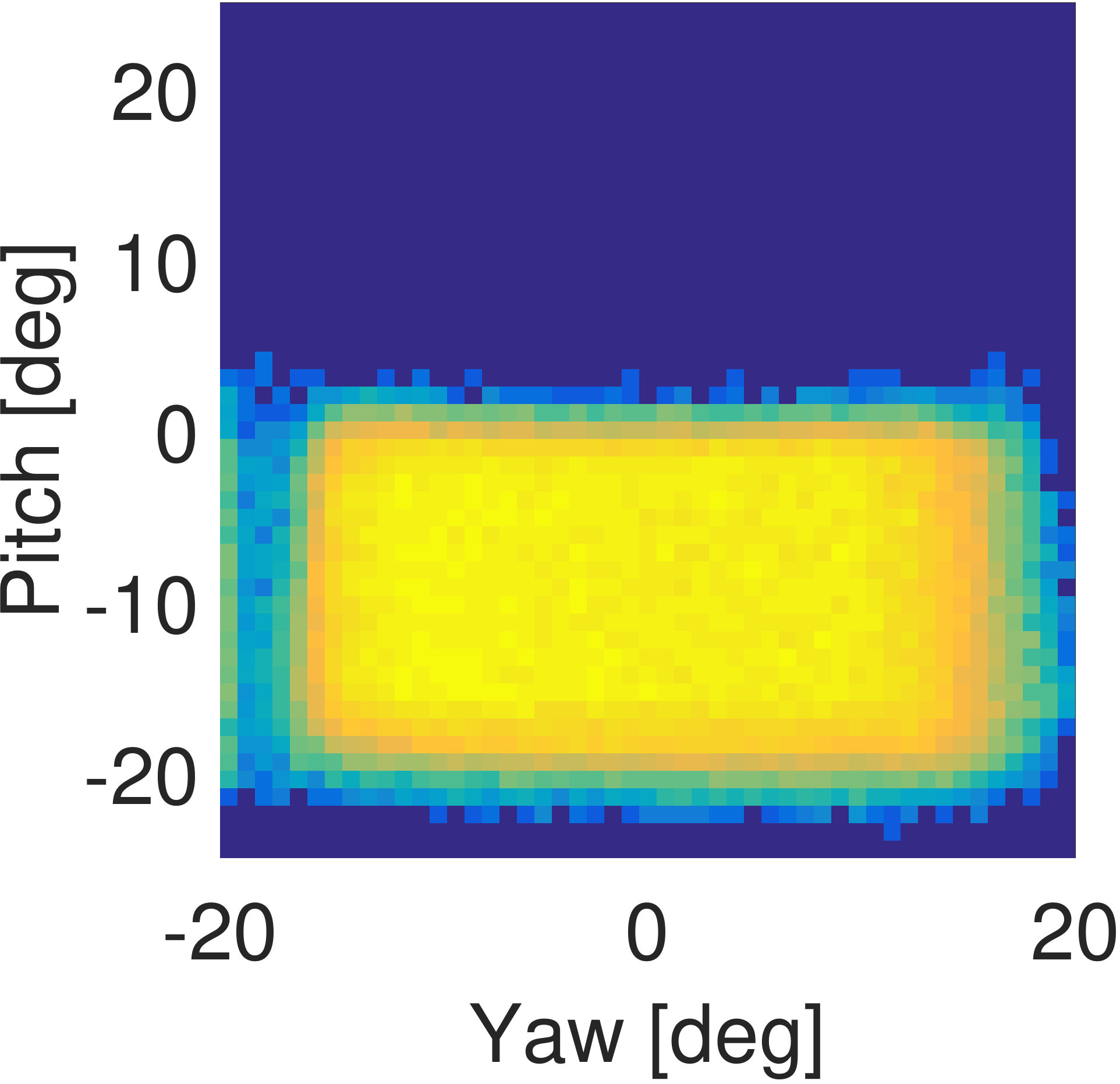}
        \caption{$\mathbf{g}$(MPIIGaze)}
    \end{subfigure}%
    \begin{subfigure}[t]{0.167\textwidth}
		\includegraphics[width=0.985\textwidth]{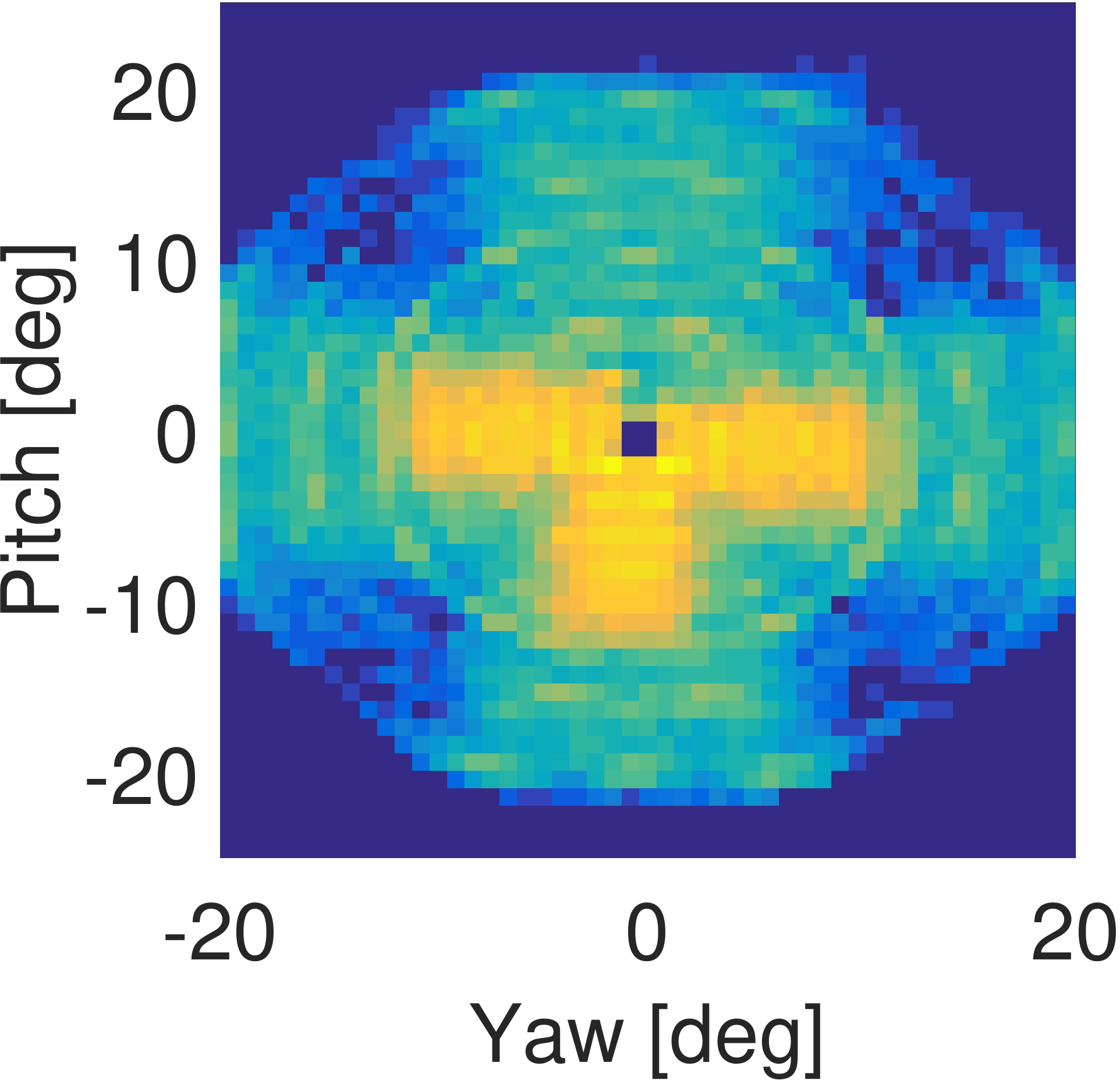}
        \caption{$\mathbf{g}$(GazeCapture)}
    \end{subfigure}%

	\caption{Distribution of head pose $\mathbf{h}$ (1\textsuperscript{st} row) and gaze direction $\mathbf{g}$ relative to the head pose (2\textsuperscript{nd} row) for datasets TabletGaze, MPIIGaze, and GazeCapture (ours). All intensities are logarithmic.}
\label{fig:DatasetDistribution}
\end{figure}

To demonstrate the variability of our data, we used the approach from~\cite{zhang15_cvpr} to estimate head pose, $\mathbf{h}$, and gaze direction, $\mathbf{g}$, for each of our frames. In Fig.~\ref{fig:DatasetDistribution} we plot the distribution of $\mathbf{h}$ and $\mathbf{g}$ on GazeCapture as well as existing state-of-the-art datasets, MPIIGaze~\cite{zhang15_cvpr} and TabletGaze~\cite{huang2015tabletgaze}. We find that while our dataset contains a similar overall distribution of $\mathbf{h}$ there is a significantly larger proportion of outliers as compared to existing datasets. Further, we observe that our data capture technique from Sec.~\ref{sec:crowdsource} introduces significant variation in the relative position of the camera to the user as compared to other datasets; \eg, we have frames where the camera is mounted below the screen (\ie, when the device is turned upside down) as well as above. These variations can be helpful for training and evaluating eye tracking approaches.

\begin{figure*}[t]
\begin{center}
   \includegraphics[width=0.8\linewidth]{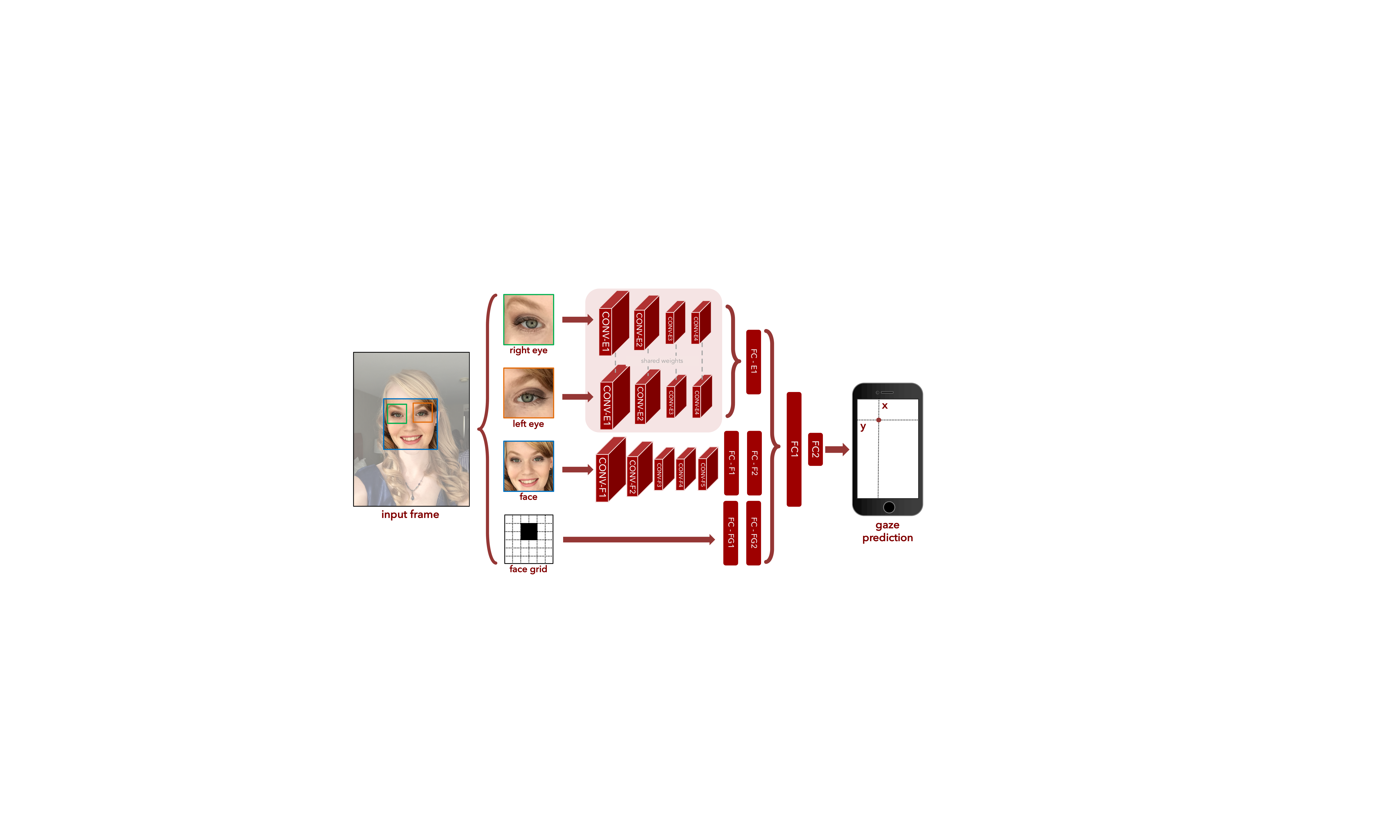}
\end{center}
\vspace*{-4mm}
\caption{Overview of iTracker, our eye tracking CNN. Inputs include left eye, right
eye, and face images detected and cropped from the original frame (all of size $224\times224$). The face grid
input is a binary mask used to indicate the location and size of the head within the frame (of size $25\times25$).
The output is the distance, in centimeters, from the camera. {\tt CONV} represents convolutional layers (with filter size/number of kernels: {\tt CONV-E1,CONV-F1}: $11\times11$/96, {\tt CONV-E2,CONV-F2}: $5\times5$/256, {\tt CONV-E3,CONV-F3}: $3\times3$/384, {\tt CONV-E4,CONV-F4}: $1\times1$/64) while {\tt FC} represents fully-connected layers (with sizes: {\tt FC-E1}: 128, {\tt FC-F1}: 128, {\tt FC-F2}: 64, {\tt FC-FG1}: 256, {\tt FC-FG2}: 128, {\tt FC1}: 128, {\tt FC2}: 2). The exact model configuration is available on the \href{http://gazecapture.csail.mit.edu}{project website}.}
\label{fig:method}
\end{figure*}

\section{iTracker: A Deep Network for Eye Tracking}
In this section, we describe our approach for building a robust eye tracker using our large-scale dataset, GazeCapture. Given the recent success of convolutional neural networks (CNNs) in computer vision, we use this approach to tackle the problem of eye tracking. We believe that, given enough data, we can learn eye tracking end-to-end without the need to include any manually engineered features, such as head pose~\cite{zhang15_cvpr}. In Sec.~\ref{sec:model}, we describe how we design an end-to-end CNN for robust eye tracking. Then, in Sec.~\ref{sec:realtime} we use the concept of \textit{dark knowledge}~\cite{hinton2015distilling} to learn a smaller network that achieves a similar performance while running at $10$--$15$fps on a modern mobile device.

\subsection{Learning an End-to-End Model}
\label{sec:model}
Our goal is to design an approach that can use the information from a single image to robustly predict gaze. We choose to use deep convolutional neural networks (CNNs) to make effective use of our large-scale dataset. Specifically, we provide the following as input to the model: (1) the image of the face together with its location in the image (termed \emph{face grid}), and (2) the image of the eyes. We believe that using the model can (1) infer the head pose relative to the camera, and (2) infer the pose of the eyes relative to the head. By combining this information, the model can infer the location of gaze. Based on this information, we design the overall architecture of our iTracker network, as shown in Fig.~\ref{fig:method}. The size of the various layers is similar to those of AlexNet~\cite{krizhevsky2012imagenet}. Note that we include the eyes as individual inputs into the network (even though the face already contains them) to provide the network with a higher resolution image of the eye to allow it to identify subtle changes.

\begin{figure}[t]
\begin{center}
   \includegraphics[width=0.65\linewidth]{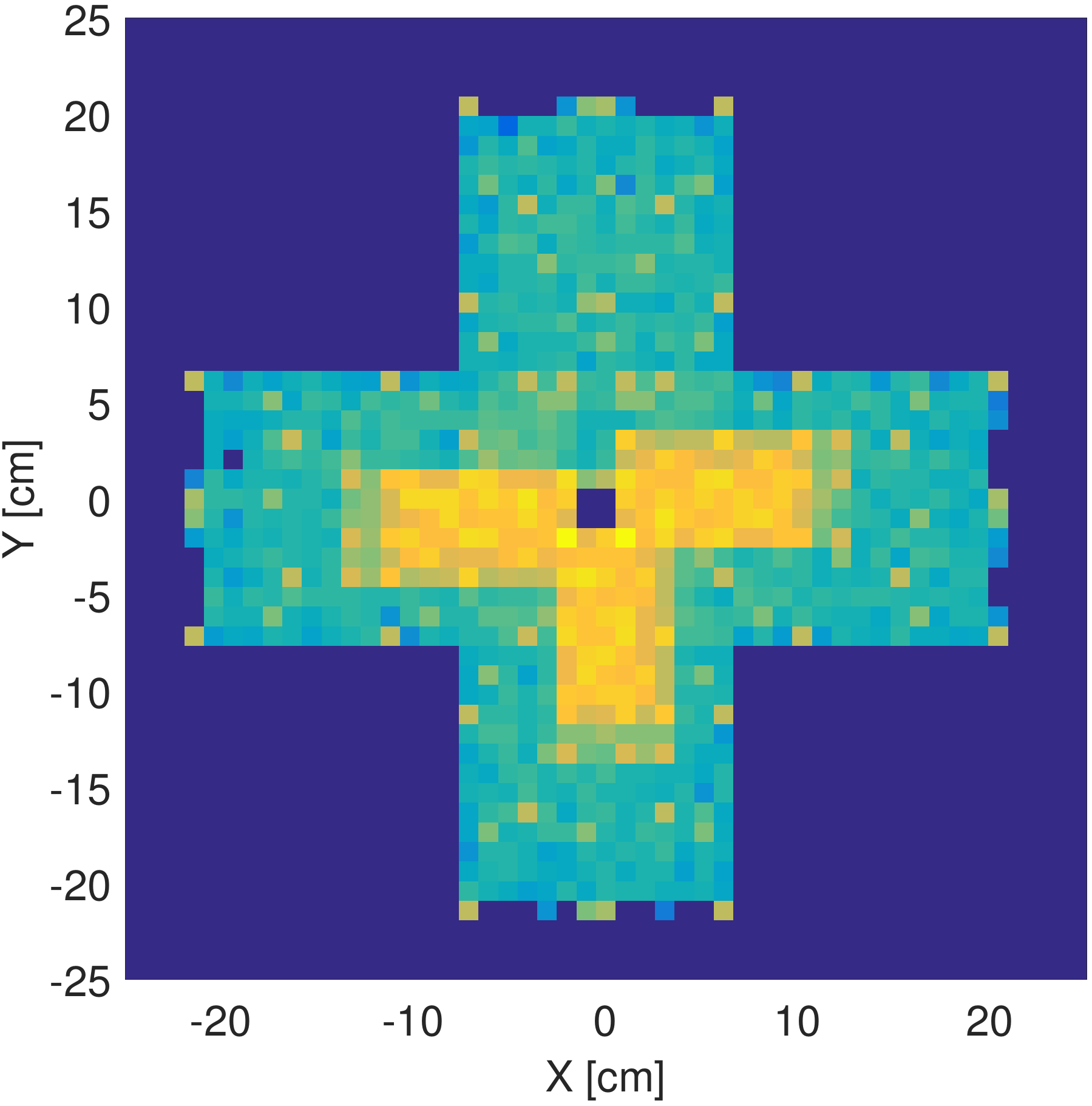}
\end{center}
\vspace*{-6mm}
\caption{Our unified prediction space. The plot above shows the distribution of all dots in our dataset mapped to the prediction space. Axes
denote centimeters from the camera; \ie, all dots on the screen are projected to this space where the camera is at $(0, 0)$.}
\label{fig:dotplot}
\end{figure}

In order to best leverage the power of our large-scale dataset, we design a unified prediction space that allows us to train a single model using all the data. Note that this is not trivial since our data was collected using multiple devices at various orientations. Directly predicting screen coordinates would not be meaningful beyond a single device in a single orientation since the input could change significantly. Instead, we leverage the fact that the front-facing camera is typically on the same plane as, and angled perpendicular to, the screen. As shown in Fig.~\ref{fig:dotplot}, we predict the dot location relative to the camera (in centimeters in the $x$ and $y$ direction). We obtain this through precise measurements of device screen sizes and camera placement. Finally, we train the model using a Euclidean loss on the $x$ and $y$ gaze position. The training parameters are provided in Sec.~\ref{sec:setup}.

Further, after training the joint network, we found fine-tuning the network to each device and orientation helpful. This was particularly useful in dealing with the unbalanced data distribution between mobile phones and tablets. We denote this model as iTracker$^*$.

\subsection{Real-Time Inference}
\label{sec:realtime}
As our goal is to build an eye tracker that is practically useful, we provide
evidence that our model can be applied on resource-constrained mobile devices.
Encouraged by the work of Hinton \etal~\cite{hinton2015distilling}, we apply
dark knowledge to reduce model complexity and thus, computation time and
memory footprint. First, while we designed the iTracker network to be robust to poor-quality eye detections, we use tighter crops (of size $80\times80$) produced by facial landmark eye detections \cite{baltrusaitis2013constrained} for the smaller network. These tighter crops focus the attention of the network on the more discriminative regions of the image, while also being faster due to the reduced image size. Then, we fine-tune the architecture configuration using the validation set to optimize efficiency without sacrificing much accuracy. Specifically, we have a combined loss on the ground truth, the predictions from our full model, as well as the features from the penultimate layer to assist the network in producing quality results. We implemented this model on an iPhone using Jetpac's Deep Belief SDK\footnote{\href{https://github.com/jetpacapp/DeepBeliefSDK}{https://github.com/jetpacapp/DeepBeliefSDK}}. We found that the reduced version of the model took about $0.05$sec. to run on a iPhone 6s. Combining this with Apple's face detection pipeline, we can expect to achieve an overall detection rate of $10$--$15$fps on a typical mobile device.

\section{Experiments}
\label{sec:experiments}

In this section, we thoroughly evaluate the performance of iTracker using our large-scale GazeCapture dataset. Overall, we significantly outperform state-of-the-art approaches, achieving an average error of $\sim2$cm without calibration and are able to reduce this further to $1.8$cm through calibration. Further, we demonstrate the importance of having a large-scale dataset as well as having variety in the data in terms of number of subjects rather than number of examples per subject. Then, we apply the features learned by iTracker to an existing dataset, TabletGaze~\cite{huang2015tabletgaze}, to demonstrate the generalization ability of our model.

\subsection{Setup}
\label{sec:setup}
\textbf{Data preparation:} First, from the 2,445,504 frames in GazeCapture, we select 1,490,959 frames that have both face and eye detections. These detections serve as important inputs to the model, as described in Sec.~\ref{sec:model}. This leads to a total of 1471 subjects being selected where each person has at least one frame with a valid detection. Then, we divide the dataset into train, validation, and test splits consisting of 1271, 50, and 150 subjects\footnote{Train, validation and test splits contain 1,251,983, 59,480, and 179,496 frames, respectively.}, respectively. For the validation and test splits, we only select subjects who looked at the full set of points. This ensures a uniform data distribution in the validation/test sets and allows us to perform a thorough evaluation on the impact of calibration across these subjects. Further, we evaluate the performance of our approach by augmenting the training and test set 25-fold by shifting the eyes and the face, changing face grid appropriately. For training, each of the augmented samples is treated independently while for testing, we average the predictions of the augmented samples to obtain the prediction on the original test sample (similar to~\cite{krizhevsky2012imagenet}).

\textbf{Implementation details:} The model was implemented using Caffe~\cite{jia2014caffe}. It was trained from scratch on the GazeCapture dataset for $150,000$ iterations with a batch size of 256. An initial learning rate of 0.001 was used, and after $75,000$ iterations, it was reduced to $0.0001$. Further, similar to AlexNet~\cite{krizhevsky2012imagenet}, we used a momentum of 0.9 and weight decay of 0.0005 throughout the training procedure. Further, we truncate the predictions based on the size of the device.

\textbf{Evaluation metric:} Similar to~\cite{huang2015tabletgaze}, we report the error in terms of average Euclidean distance (in centimeters) from the location of the true fixation. Further, given the different screen sizes, and hence usage distances of phones and tablets, we provide performance for both of these devices (even though the models used are exactly the same for both devices, unless otherwise specified). Lastly, to simulate a realistic use case where a stream of frames is processed for each given fixation rather than just a single frame, we report a value called \textit{dot error}. In this case, the output of the classifier is given as the average prediction of all the frames corresponding to a gaze point at a certain location.

\subsection{Unconstrained Eye Tracking}
\begin{figure}[t]
\begin{center}
   \includegraphics[width=1\linewidth]{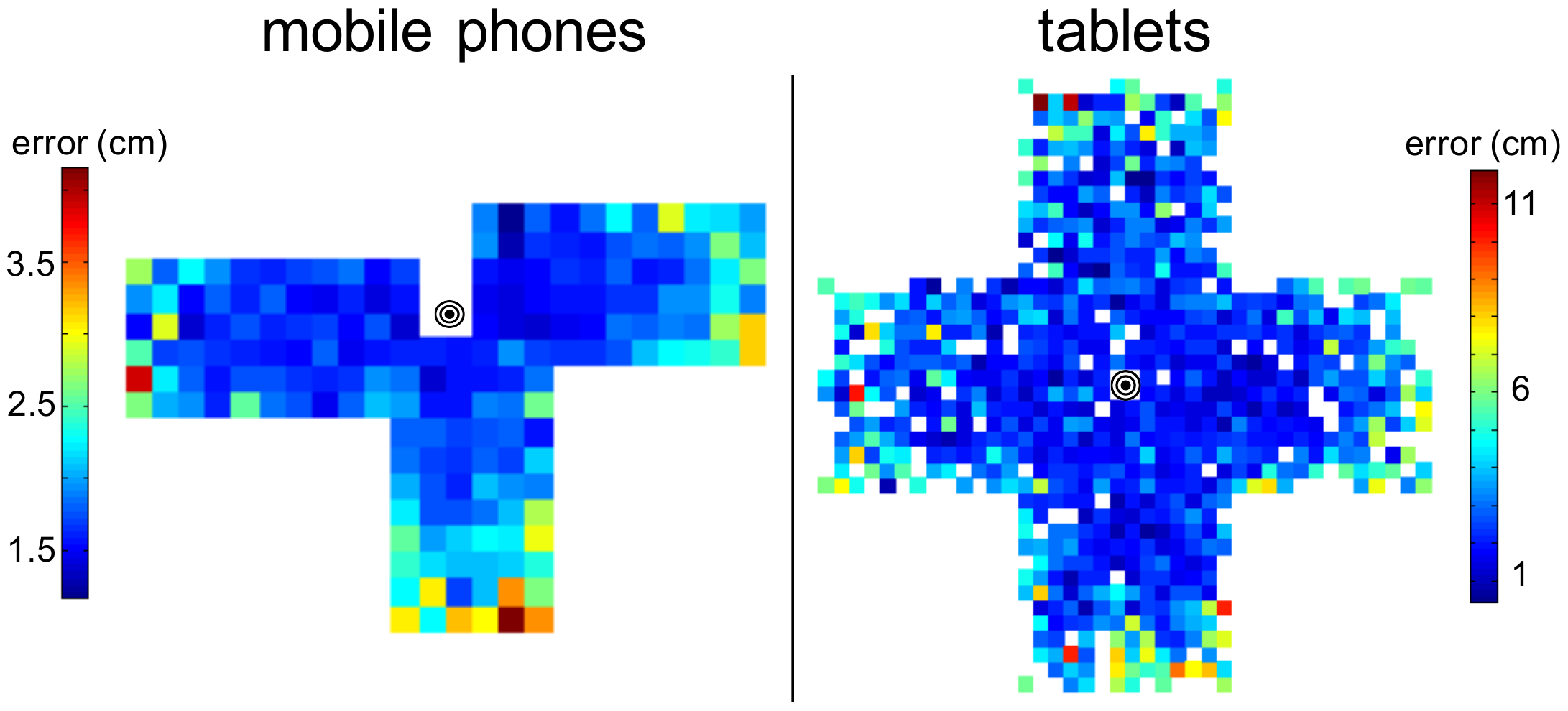}
\end{center}
\vspace*{-6mm}
\caption{Distribution of error for iTracker (with train and test augmentation) across the prediction space, plotted at ground truth location. The black and white circles represent the location of the camera. We observe that the error near the camera tends to be lower.}
\label{fig:heatmap}
\end{figure}

Here, our goal is to evaluate the generalization ability of iTracker to novel faces by evaluating it on unconstrained (calibration-free) eye tracking. As described in Sec.~\ref{sec:setup}, we train and test iTracker on the appropriate splits of the data. To demonstrate the impact of performing data augmentation during train and test, we include the performance with and without train/test augmentation. As baseline, we apply the best performing approach (pre-trained ImageNet model) on TabletGaze (Sec.~\ref{sec:tabletgaze}) to GazeCapture. The results are summarized in the top half of Tbl.~\ref{tbl:calibrationfree} and the error distribution is plotted in Fig.~\ref{fig:heatmap}.

We observe that our model consistently outperforms the baseline approach by a large margin, achieving an error as low as 1.53cm and 2.38cm on mobile phones and tablets respectively. Further, we find that the \textit{dot error} is consistently lower than the \textit{error} demonstrating the advantage of using temporal averaging in real-world eye tracking applications. Also note that both train and test augmentation are helpful for reducing the prediction error. While test augmentation may not allow for real-time performance, train augmentation can be used to learn a more robust model. Last, we observe that fine-tuning the general iTracker model to each device and orientation (iTracker$^*$) is helpful for further reducing errors, especially for tablets. This is to be expected, given the large proportion of samples from mobile phones (85\%) as compared to tablets (15\%) in GazeCapture.

\begin{table}
 \centering
  \small
 \begin{tabular}{l|c|@{\hspace*{2mm}}cc@{}|c@{\hspace*{4mm}}c@{\hspace*{1mm}}}
 \hline
\multirow{2}{*}{Model} & \multirow{2}{*}{Aug.} & \multicolumn{2}{c|}{Mobile phone} & \multicolumn{2}{c}{Tablet} \\
   & & error & dot err. & error & dot err.\\\hline\hline
  Baseline & tr $+$ te & 2.99 & 2.40 & 5.13 & 4.54\\\hline\hline
  iTracker & None & 2.04 & 1.62 & 3.32 & 2.82 \\\hline
  iTracker & te & 1.84 & 1.58 & 3.21 & 2.90 \\\hline
  iTracker & tr & 1.86 & 1.57 & 2.81 & 2.47 \\\hline
  iTracker & tr $+$ te & 1.77 & \textbf{1.53} & 2.83 & 2.53\\\hline
  iTracker$^*$ & tr $+$ te & \textbf{1.71} & \textbf{1.53} & \textbf{2.53} & \textbf{2.38} \\\hline\hline
  iTracker (no eyes) & None & 2.11 & 1.72 & 3.40 & 2.93 \\\hline
  iTracker (no face) & None & 2.15 & 1.69 & 3.45 & 2.92 \\\hline
  iTracker (no fg.) & None & 2.23 & 1.81 & 3.90 & 3.36 \\\hline
 \end{tabular}
 \vspace*{-2mm}
 \caption{Unconstrained eye tracking results (top half) and ablation study (bottom half). The error and dot error values are reported in centimeters (see Sec.~\ref{sec:setup} for details); lower is better. \textit{Aug.} refers to dataset augmentation, and \textit{tr} and \textit{te} refer to train and test respectively. \textit{Baseline} refers to applying support vector regression (SVR) on features from a pre-trained ImageNet network, as done in Sec.~\ref{sec:tabletgaze}. We found that this method outperformed all existing approaches. For the ablation study (Sec.~\ref{sec:analysis}), we removed each critical input to our model, namely eyes, face and face grid (\textit{fg.}), one at a time and evaluated its performance.}
 \label{tbl:calibrationfree}
\end{table}

\subsection{Eye Tracking with Calibration}
\label{sec:calibration}
As mentioned in Sec.~\ref{sec:crowdsource}, we collect data from 13 fixed dot locations (per device orientation) for each subject. We use these locations to simulate the process of calibration. For each subject in the test set, we use frames from these 13 fixed locations for training, and evaluate on the remaining locations. Specifically, we extract features from the {\tt fc1} layer of iTracker and train a model using SVR to predict each subject's gaze locations. The results are summarized in Tbl.~\ref{tbl:calibration}. We observe that the performance decreases slightly when given few points for calibration. This likely occurs due to overfitting when training the SVR. However, when using the full set of 13 points for calibration, the performance improves significantly, achieving an error of 1.34cm and 2.12cm on mobile phones and tablets, respectively.

\begin{table}
 \centering
  \small
 \begin{tabular}{l|c|cc|cc}
 \hline
  \multirow{2}{*}{Model} & \# calib. & \multicolumn{2}{c|}{Mobile phone} & \multicolumn{2}{c}{Tablet} \\
   & points & error & dot err. & error & dot err.\\\hline\hline
  \multirow{5}{*}{iTracker} & 0 & 1.77 & 1.53 & 2.83 & 2.53 \\
 & 4 & 1.92 & 1.71 & 4.41 & 4.11 \\
& 5 & 1.76 & 1.50 & 3.50 & 3.13 \\
& 9 & 1.64 & 1.33 & 3.04 & 2.59 \\
& 13 & \textbf{1.56} & \textbf{1.26} & \textbf{2.81} & \textbf{2.38}\\\hline\hline
  \multirow{5}{*}{iTracker*} & 0 & 1.71 & 1.53 & 2.53 & 2.38 \\
 & 4 & 1.65 & 1.42 & 3.12 & 2.96 \\
& 5 & 1.52 & 1.22 & 2.56 & 2.30 \\
& 9 & 1.41 & 1.10 & 2.29 & 1.87 \\
& 13 & \textbf{1.34} & \textbf{1.04} & \textbf{2.12} & \textbf{1.69} \\\hline
 \end{tabular}
 \caption{Performance of iTracker using different numbers of points for calibration (error and dot error in centimeters; lower is better). Calibration significantly improves performance.}
 \label{tbl:calibration}
\end{table}

\subsection{Cross-Dataset Generalization}
\label{sec:tabletgaze}
We evaluate the generalization ability of the features learned by iTracker by applying them to another dataset, TabletGaze~\cite{huang2015tabletgaze}. TabletGaze contains recordings from a total of 51 subjects and a sub-dataset of 40 usable subjects\footnote{~\cite{huang2015tabletgaze} mentions 41 usable subjects but at the time of the experiments, only data from 40 of them was released.}. We split this set of 40 subjects into 32 for training and 8 for testing. We apply support vector regression (SVR) to the features extracted using iTracker to predict the gaze locations in this dataset, and apply this trained classifier to the test set. The results are shown in Tbl.~\ref{tbl:baselines}. We report the performance of applying various state-of-the-art approaches (TabletGaze~\cite{huang2015tabletgaze}, TurkerGaze~\cite{xu2015turkergaze} and MPIIGaze~\cite{zhang15_cvpr}) and other baseline methods for comparison. We propose two simple baseline methods: (1) center prediction (\ie, always predicting the center of the screen regardless of the data) and (2) applying support vector regression (SVR) to image features extracted using AlexNet~\cite{krizhevsky2012imagenet} pre-trained on ImageNet~\cite{russakovsky2014imagenet}. Interestingly, we find that the AlexNet + SVR approach outperforms all existing state-of-the-art approaches despite the features being trained for a completely different task. Importantly, we find that the features from iTracker significantly outperform all existing approaches to achieve an error of 2.58cm demonstrating the generalization ability of our features.

\begin{table}
 \centering
  \small
 \begin{tabular}{l|c|c}
 \hline
  Method & Error & Description\\\hline\hline
  Center & 7.54 & Simple baseline \\\hline
  TurkerGaze~\cite{xu2015turkergaze} & 4.77 & pixel features + SVR \\\hline
  TabletGaze & 4.04 & Our implementation of ~\cite{huang2015tabletgaze}\\\hline
  MPIIGaze~\cite{zhang15_cvpr} & 3.63 & CNN + head pose\\\hline
  TabletGaze\cite{huang2015tabletgaze}& 3.17 & Random forest $+$ mHoG\\\hline
  AlexNet~\cite{krizhevsky2012imagenet} & 3.09 & eyes ({\tt conv3}) $+$ face ({\tt fc6}) $+$ fg.\\\hline\hline
  iTracker (ours) & \textbf{2.58} & {\tt fc1} of iTracker + SVR \\\hline
 \end{tabular}
 \vspace*{-3mm}
 \caption{Result of applying various state-of-the-art approaches to TabletGaze~\cite{huang2015tabletgaze} dataset (error in cm). For the AlexNet + SVR approach, we train a SVR on the concatenation of features from various layers of AlexNet ({\tt conv3} for eyes and {\tt fc6} for face) and a binary face grid (\textit{fg.}).}
 \vspace*{-2mm}
 \label{tbl:baselines}
\end{table}

\subsection{Analysis}
\label{sec:analysis}
\textbf{Ablation study:} In the bottom half of Tbl.~\ref{tbl:calibrationfree} we report the performance after removing different components of our model, one at a time, to better understand their significance. In general, all three inputs (eyes, face, and face grid) contribute to the performance of our model. Interestingly, the mode with face but no eyes achieves comparable performance to our full model suggesting that we may be able to design a more efficient approach that requires only the face and face grid as input. We believe the large-scale data allows the CNN to effectively identify the fine-grained differences across people's faces (their eyes) and hence make accurate predictions.

\begin{figure}[t]
    \centering
        \begin{subfigure}[t]{0.46\linewidth}
		\includegraphics[width=\textwidth]{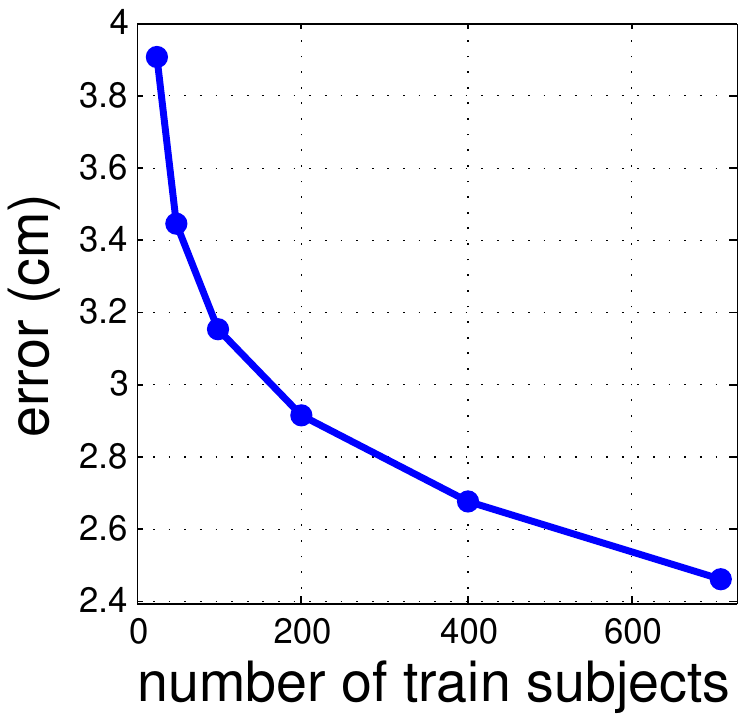}
        \caption{No. of subjects \vs error}
    \end{subfigure}%
    \hspace*{2mm}
    \begin{subfigure}[t]{0.49\linewidth}
      \includegraphics[width=\textwidth]{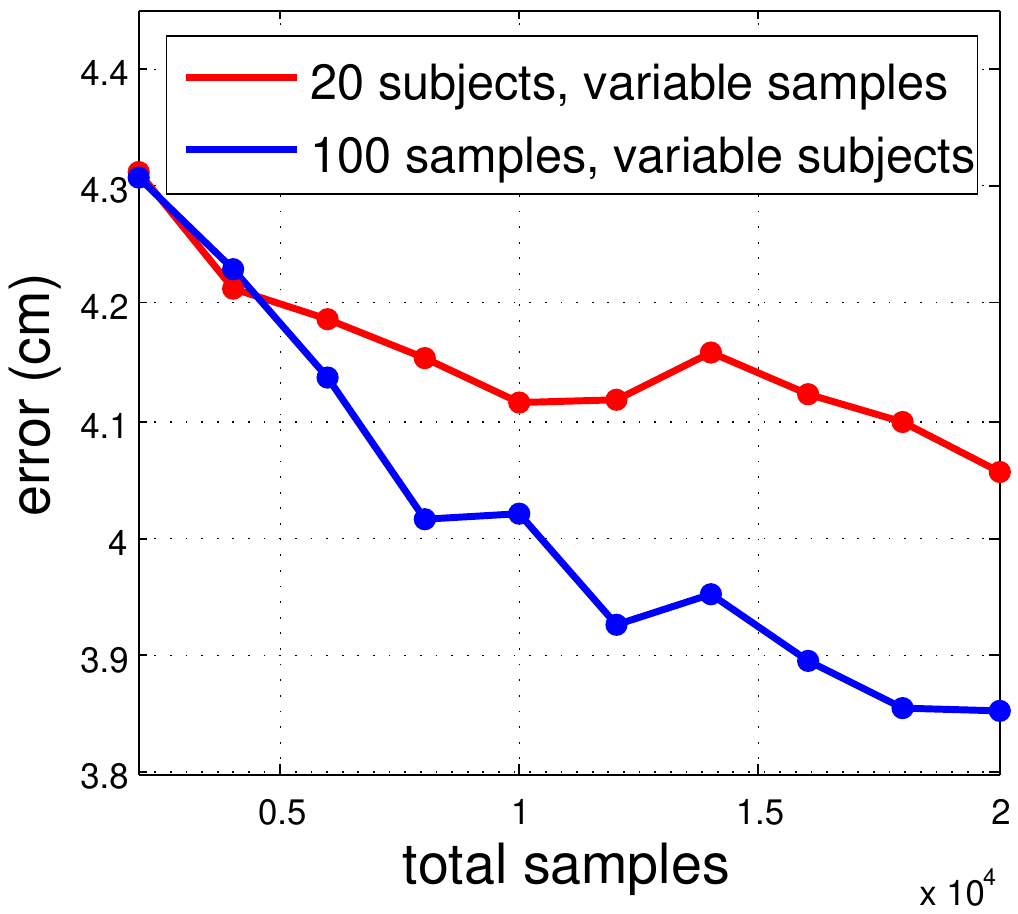}
      \caption{Subjects \vs samples}
       \label{fig:subjects_vs_samples}
    \end{subfigure}%
\vspace*{-2mm}
    \caption{Dataset size is important for achieving low error. Specifically, growing the number of subjects in a dataset is more important than the number of samples, which further motivates the use of crowdsourcing.}
    \label{fig:datasetsize}
\end{figure}

\textbf{Importance of large-scale data:} In Fig.~\ref{fig:subjects_vs_samples} we plot the performance of iTracker as we increase the total number of train subjects. We find that the error decreases significantly as the number of subjects is increased, illustrating the importance of gathering a large-scale dataset. Further, to illustrate the importance of having variability in the data, in Fig.~\ref{fig:subjects_vs_samples}, we plot the performance of iTracker as (1) the number of subjects is increased while keeping the number of samples per subject constant (in blue), and (2) the number of samples per subject is increased while keeping the number of subjects constant (in red). In both cases the total number of samples is kept constant to ensure the results are comparable. We find that the error decreases significantly more quickly as the number of subjects is increased indicating the importance of having variability in the data.

\section{Conclusion}

In this work, we introduced an end-to-end eye tracking solution targeting mobile devices. First, we introduced GazeCapture, the first large-scale mobile eye tracking dataset. We demonstrated the power of crowdsourcing to collect gaze data, a method unexplored by prior works. We demonstrated the importance of both having a large-scale dataset, as well as having a large variety of data to be able to train robust models for eye tracking. Then, using GazeCapture we trained iTracker, a deep convolutional neural network for predicting gaze. Through careful evaluation, we show that iTracker is capable of robustly predicting gaze, achieving an error as low as 1.04cm and 1.69cm on mobile phones and tablets respectively. Further, we demonstrate that the features learned by our model generalize well to existing datasets, outperforming  state-of-the-art approaches by a large margin. Though eye tracking has been around for centuries, we believe that this work will serve as a key benchmark for the next generation of eye tracking solutions. We hope that through this work, we can bring the power of eye tracking to everyone.

\section*{Acknowledgements}
We would like to thank Kyle Johnsen for his help with the IRB, as well as Bradley Barnes and Karen Aguar for helping to recruit participants. This research was supported by Samsung, Toyota, and the QCRI-CSAIL partnership.

{\footnotesize
\bibliographystyle{ieee}
\bibliography{itracker}
}

\end{document}